\documentclass{ceurart}

\usepackage{acro}
\usepackage[british]{babel}
\usepackage{booktabs}
\usepackage{cleveref}
\usepackage{csquotes}
\usepackage[inline]{enumitem}
\usepackage[shortcuts]{extdash}
\usepackage{listings}
\usepackage{microtype}
\usepackage[all]{nowidow}
\usepackage{pgfplots}
\usepackage{siunitx}
\usepackage{tikz}
\usepackage[most]{tcolorbox}
\usepackage{upgreek}
\usepackage{xcolor}

\DeclareAcronym{am}{
  long=any match,
  short=AM,
}
\DeclareAcronym{cb}{
  long=closed-book,
  short=CB,
}
\DeclareAcronym{ci}{
  long=confidence interval,
  short=CI,
}
\DeclareAcronym{dyprag}{
  long=dynamic parametric RAG,
  short=DyPRAG,
}
\DeclareAcronym{em}{
  long=exact match,
  short=EM,
}
\DeclareAcronym{kg}{
  long=knowledge graph,
  short=KG,
}
\DeclareAcronym{kgqa}{
  long=knowledge graph question answering,
  short=\textsc{kgqa},
}
\DeclareAcronym{llm}{
  long=large language model,
  short=LLM,
}
\DeclareAcronym{lora}{
  long=low-rank adaptation,
  short=LoRA,
}
\DeclareAcronym{ob}{
  long=open-book,
  short=OB,
}
\DeclareAcronym{qa}{
  long=question answering,
  short=QA,
}
\DeclareAcronym{rag}{
  long=retrieval-augmented generation,
  short=RAG,
}
\DeclareAcronym{svd}{
  long=singular value decomposition,
  short=SVD,
}

\DeclareSIUnit{\GPU}{GPU}
\newtcblisting{prompt}{
  breakable,
  listing only,
  listing options={
    basicstyle=\ttfamily,
    breakatwhitespace=true,
    breakautoindent=false,
    breakindent=0pt,
    breaklines=true,
    columns=fullflexible,
    keepspaces=true,
    postbreak={\raisebox{0ex}[0ex][0ex]{\ensuremath{\hookrightarrow\space}}},
  }
}

\usetikzlibrary{
  arrows.meta,
  positioning,
}

\definecolor{my_blue}{HTML}{15699A}
\definecolor{my_darkgray}{HTML}{595959}
\definecolor{my_gray}{HTML}{707070}
\definecolor{my_lightgray}{HTML}{B8B8B8}

\begin{document}

\copyrightyear{2026}
\copyrightclause{
  Copyright for this paper by its authors.
  Use permitted under Creative Commons License Attribution 4.0 International (CC BY 4.0).
}

\conference{SKGi 2026: Scaling Knowledge Graphs for Industry Workshop, co-located with SEMANTiCS'26: 22nd International Conference on Semantic Systems, September 15--17, 2026, Ghent, Belgium}

\title{A Storage--Retrieval Gap in Parametric Knowledge Graph~Memory}

\author[1,2]{Martino M. L. Pulici}[
  email=martino.pulici@de.bosch.com,
  orcid=0009-0009-4533-7495,
  url=https://ilsommo.github.io,
]
\cormark[1]
\address[1]{Bosch Center for Artificial Intelligence, Robert-Bosch-Campus 1, 71272 Renningen, Germany}
\address[2]{LMU Munich, Institute for Informatics, Oettingenstra{\ss}e 67, 80538 Munich, Germany}
\author[1]{Cuong Xuan Chu}[
  email=cuongxuan.chu@de.bosch.com,
  url=https://cuongcx.github.io,
]
\author[1,3]{Evgeny Kharlamov}[
  email=evgeny.kharlamov@de.bosch.com,
  orcid=0000-0003-3247-4166,
  url=https://www.bosch.com/research/about-bosch-research/our-research-experts/evgeny-kharlamov,
]
\address[3]{University of Oslo, Department of Informatics, Gaustadall{\'e}en 23 B, 0373 Oslo, Norway}
\author[2,4]{Volker Tresp}[
  email=volker.tresp@lmu.de,
  orcid=0000-0001-9428-3686,
  url=https://www.dbs.ifi.lmu.de/~tresp,
]
\address[4]{Munich Center for Machine Learning, LMU Munich, Institute for Informatics, Oettingenstra{\ss}e 67, 80538 Munich, Germany}
\cortext[1]{Corresponding author.}

\begin{abstract}
  Graph retrieval-augmented generation places retrieved subgraphs into the model's context window at query time, paying a recurring token cost and exposing source data on every call.
  We study an alternative: compiling a knowledge graph offline into a bank of LoRA adapters, one per entity, that serve as a parametric knowledge layer queried by injecting weights rather than text, at zero query-time context cost.
  On the MetaQA dataset, we find that subgraph-trained adapters encode context-free factual knowledge that generalizes to unseen questions: on single-valued relations the adapter gains \num{+0.243} exact-match score over a base model that is nearly blind closed-book (\num{0.007}), and only the correct adapter recovers this knowledge (an oracle gap of \num{+0.283} over the base model).
  However, the stored knowledge is not recoverable by similarity: given a query with no subgraph, embedding-based and weight-space geometry retrieval both perform at chance, because a semantically neighbouring entity's adapter does not contain the answer\---knowledge is stored locally and does not transfer.
  Weight geometry correlates with subgraph semantics (\(\rho = \num{+0.329}\)) but not with functional retrievability.
  We quantify the byte and context-token costs against graph retrieval-augmented generation and discuss deployment implications.
  Our results establish that parametric knowledge graph memory is feasible for storing knowledge, and identify selecting and composing the right adapters by a mechanism other than semantic similarity as the central open problem\---motivating a learned, query-conditioned composition mechanism.
\end{abstract}


\maketitle

\section{Introduction}

\Acp{kg} give industrial AI systems a structured, auditable substrate of facts, and a now-standard way to put that substrate to work with \acp{llm} is graph \ac{rag}: retrieve a relevant subgraph for a query, serialize it into the prompt, and let the model read the answer out of context.
This pattern is effective but carries two recurring costs that scale with usage rather than with the size of the graph.
First, every query pays a token cost: the serialized subgraph consumes context-window budget, inflating latency, memory, and energy.
Second, every query incurs a data-exposure cost: the raw subgraph is shipped to wherever inference runs, which is problematic when the graph encodes sensitive or proprietary engineering knowledge.

We investigate an alternative in which the cost is paid once, offline.
A \ac{kg} is compiled, entity neighbourhood by entity neighbourhood, into a bank of \ac{lora} adapters~\citep{hu2022lora}.
Each adapter is a small weight perturbation \(\mathbf{\Delta W}\) that encodes one subgraph.
At query time the system retrieves one or more adapters, injects them into a frozen base model, and answers with no subgraph text in the context window.
The knowledge graph thus becomes a parametric memory: a set of weights, not a corpus of text.

Whether this works at all turns on a question that prior parametric \ac{rag} systems leave largely implicit: when an adapter is fine-tuned on a piece of text, does it store that text's facts in its weights, or does it only learn to use the text when the text is present at inference? The distinction is decisive for parametric memory, because at query time the model has no subgraph in context\---so only knowledge that resides in the weights is available at all.
We therefore ask three questions:
\begin{enumerate}
  \item Do subgraph adapters actually store knowledge recoverable with no context?
  \item Can a functional adapter\---one that answers the query\---be retrieved from the query alone?
  \item Can adapters be composed across entities for multi-hop questions?
\end{enumerate}
Our findings separate cleanly.
Subgraph adapters do store generalizable closed-book knowledge: on a compact model and a near-blind base, single-valued relations gain \num{+0.243} \ac{em}, and only the entity's own adapter recovers it (a \num{+0.283} gain over the base model).
However, the stored knowledge turns out to be local and non-retrievable: given a query with no subgraph, neither question-embedding nor weight-space geometry finds an adapter that answers it above chance, because a semantically similar entity's adapter simply does not contain the answer.
Our central finding is therefore a dissociation.
A \ac{kg} can be compiled into per-entity \ac{lora} adapters that store context-free, generalizable knowledge at zero query-time token cost.
Yet, selecting and composing the right adapters for a query is unsolved: knowledge is stored locally, and similarity-based retrieval does not recover it.

We make the following contributions:
\begin{itemize}
  \item a parametric \ac{kg} memory formulation in which per-entity subgraphs are compiled into a \ac{lora} adapter bank, queried by weight injection with zero subgraph tokens at inference (\cref{sec:method})
  \item evidence that subgraph adapters store generalizable knowledge, since on unseen questions single-valued relations gain \num{+0.243} \ac{em} over a near-blind base and only the correct adapter recovers it (\num{+0.283} oracle gap; \cref{sec:encoding})
  \item a storage--retrievability dissociation, whereby the stored knowledge is not recoverable from a query, as embedding and \(\mathbf{\Delta W}\) geometry retrieval both perform at chance closed-book because a semantically neighbouring adapter does not contain the answer, while weight geometry correlates with semantics but not with function (\cref{sec:retrieval})
  \item a cost analysis and a precise open problem, as we quantify the zero-context-token advantage over graph \ac{rag} (\cref{sec:cost}) and show that both retrieval and multi-hop composition reduce to a single unsolved task\---learning which local adapters to combine for a query (\cref{sec:composition}).
\end{itemize}

\section{Method}\label{sec:method}

\begin{figure}
  \centering

  \begin{tikzpicture}[
      ar/.style={
        -{Latex[length=1.6mm]},
        thick,
      },
      box/.style={
        align=center,
        draw,
        fill=white,
        minimum width=2.6cm,
        rounded corners,
      },
      node distance=4mm and 6mm,
    ]
    \node[box] (sub) {
      Subgraph \\
      \(S_i\)
    };
    \node[
      box,
      right=of sub,
    ] (verb) {
      Verbalize \\
      \(\operatorname{v}(S_i)\)
    };
    \node[
      box,
      right=of verb,
    ] (train) {
      Train LoRA \\
      \(\boldsymbol{\Delta \uptheta}_i\)
    };
    \node[
      box,
      right=of train,
    ] (bank) {
      Adapter bank \\
      \(\mathcal{B}\)
    };
    \draw[ar] (sub) -- (verb);
    \draw[ar] (verb) -- (train);
    \draw[ar] (train) -- (bank);
    \node[
      left=1mm of sub,
      anchor=south,
      rotate=90,
      my_darkgray,
    ] {Offline};
    \node[
      below=8mm of sub,
      box,
    ] (q) {
      Query \\
      \(q\)
    };
    \node[
      box,
      right=of q,
    ] (ret) {
      Select adapters \\
      \(\boldsymbol{\Delta \uptheta}_{i} \in \mathcal{B}\)
    };
    \node[
      box,
      right=of ret,
    ] (inj) {
      Inject \\
      \(\boldsymbol{\uptheta}_0 + \sum_i \lambda_i \boldsymbol{\Delta \uptheta}_i\)
    };
    \node[
      box,
      right=of inj,
    ] (ans) {
      Answer \\
      \(a\)
    };
    \draw[ar] (q) -- (ret);
    \draw[ar] (ret) -- (inj);
    \draw[ar] (inj) -- (ans);
    \draw[ar] (bank) -- (ret);
    \node[
      left=1mm of q,
      anchor=south,
      rotate=90,
      my_darkgray,
    ] {Online};
  \end{tikzpicture}

  \caption{
    Parametric \acs{kg} memory.
    In the offline phase, each entity subgraph is verbalized and compiled into a per-entity \acs{lora} adapter, forming a bank; in the online phase, an adapter is selected for the query and injected into the frozen base, and the model answers with no subgraph in its context.
  }\label{fig:pipeline}
\end{figure}
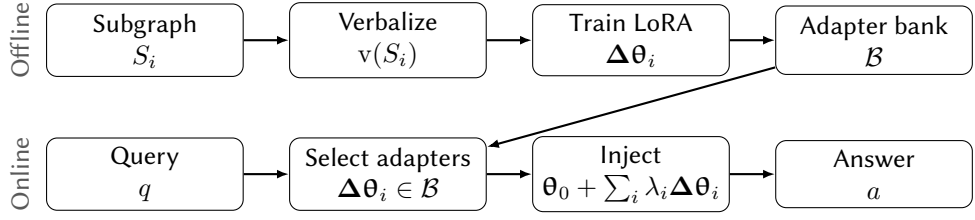

\paragraph{Parametric knowledge graph memory.}
\Cref{fig:pipeline} summarizes the design.
Offline, each entity's subgraph is verbalized and compiled into a per-entity \ac{lora} adapter, forming a bank; online, an adapter is selected for the query and injected into the frozen base, and the model answers with no subgraph in its context.
Let
\begin{equation*}
  G = \{(h, r, t) \mid h, t \in E,\ r \in R\}
\end{equation*}
be a \ac{kg} with entities \(E\) and relations \(R\).
For a chosen set of anchor entities \(\{e_1, \dots, e_N\} \subseteq E\) we extract for each \(e_i\) its \(k\)-hop neighbourhood subgraph \(S_i \subseteq G\).
Each subgraph is verbalized by a function into text as \(\operatorname{v}(S_i)\) and used to fine-tune a \ac{lora} adapter \(\boldsymbol{\Delta \uptheta}_i\) on a frozen base model \(\boldsymbol{\uptheta}_0\), producing a bank \(\mathcal{B} = {\{\boldsymbol{\Delta \uptheta}_i\}}_{i=1}^N\).
At query time, given a question, we:
\begin{enumerate*}[label=(\emph{\alph*})]
  \item retrieve one or more adapters \(\boldsymbol{\Delta \uptheta}_{i} \in \mathcal{B}\);
  \item form \(\boldsymbol{\uptheta} = \boldsymbol{\uptheta}_0 + \sum_i \lambda_i \boldsymbol{\Delta \uptheta}_i\); and
  \item decode the answer from \(\boldsymbol{\uptheta}\) with no subgraph text in its context.
\end{enumerate*}
Single-entity questions use a single adapter; multi-entity questions require merging (\cref{sec:composition}).

\paragraph{Verbalization.}
Each subgraph is rendered into text by a verbalization function \(\operatorname{v}(\cdot)\) before training.
We use a fixed, fact-dense template that enumerates the entity's triples with minimal surface variation (one declarative clause per triple), so that the adapter is trained on a compact, consistent rendering of the subgraph.
Because a \ac{kg} subgraph is already a short enumeration of facts, this templated form stays close to any natural-language rendering of the same triples; we did not find weight-level storage to depend on the choice, and use the templated function throughout.
The query-time model sees no subgraph, so only weight-level knowledge is available; that such knowledge is present at all is what makes parametric \ac{kg} memory possible.

\paragraph{The \(\mathbf{\Delta W}\) representation.}
Each adapter's perturbation \(\mathbf{\Delta W} = \mathbf{B} \mathbf{A}\) per layer--module is stored via its truncated \ac{svd}, and pairwise distances are computed as normalized Frobenius distances directly from the \ac{svd} factors without materializing \(\mathbf{\Delta W}\) (details in~\cref{sec:math}).
For retrieval (\cref{sec:retrieval}) we score a query against the bank and return the nearest adapters under \(\mathbf{\Delta W}\) geometry, contrasting this with a sentence-embedding baseline and random selection.
Critically, in deployment the query carries no subgraph, so the retrieval signal must come from the question text~alone.

\section{Experimental setup}

\paragraph{Training.}
We use Qwen3.5-2B~\citep{qwenteam2026qwen35,qwen2026qwen35}, a compact model inexpensive to train yet able to answer the benchmark's questions when the relevant facts are available.
All adapters share this frozen base, loaded in \num{4}-bit QLoRA~\citep{dettmers2023qlora}, with a standard \ac{lora} configuration (\(r = \num{8}\), \(\alpha = \num{16}\), \num{0.05} dropout, all seven projection types) trained to convergence with AdamW~\citep{loshchilov2019decoupled} (learning rate \num{2e-4}, cosine schedule, \num{50} warmup steps, batch size of four) using a fixed instruction template (\cref{app:prompts}).
Of each entity's \ac{qa} pairs, eight are used for training and four held out for evaluation.

\paragraph{Corpus.}
We use MetaQA~\citep{zhang2018variational}, a movie knowledge graph with entity-level \ac{qa}.
MetaQA has no unique film identifiers, so the film title is necessarily the entity key; because distinct films can share a title, keying naively on the title would merge them into a single adapter trained on contradictory labels.
We therefore retain only films whose title maps to exactly one release year in the knowledge base, excluding multi-film title collisions; adapters built from collided titles often fail to reproduce even their own training answers, since the same question maps to several conflicting gold answers, whereas clean single-film adapters do so reliably.
This yields \num{150} clean single-film entities, each with an internally consistent one-hop subgraph verbalized into templated triple-enumeration text.

\paragraph{Probes and metrics.}
We evaluate each adapter \ac{cb}, with no subgraph in the context, and \ac{ob}, with the subgraph in the context.
MetaQA relations are set-valued (a film has many tags, several cast members), so we report three metrics.
Our primary metric is \textbf{\ac{em}}: since relations are set-valued, a prediction counts as correct if, under normalized \ac{em}, it equals any valid object of the queried entity--relation pair.
\Ac{em} cannot score high-cardinality relations when the model emits a valid answer inside verbose output, so we additionally report \acl{am} (\textbf{\acs{am}}), measuring if the prediction contains any valid object, and \textbf{set recall}, computed against the union of valid objects.
We distinguish single-valued relations (director, writer, release year), where \ac{em} is meaningful, from multi-valued relations (tags, cast), where set metrics are required.
All metrics are applied identically to the base model and every adapter.
Generations use greedy decoding with a \mbox{\num{128}-token} cap.
All gains are reported as per-entity paired differences over base with \qty{95}{\percent} bootstrap \acp{ci} based on \num{50000}~resamples.

\section{Experiments}

\subsection{Weight-level storage}\label{sec:encoding}

Our first question is whether a subgraph-trained adapter stores knowledge that is recoverable \ac{cb}\---with no subgraph in context\---and that generalizes to questions the adapter was not trained on.
This is the defining property of a parametric memory: at query time the model sees only the question, not the~subgraph.

\paragraph{Closed-book generalization gain.}
We evaluate each adapter \ac{cb} on its entity's held-out questions and compare to the base model on the same questions.
The base model is nearly blind on \ac{cb} questions: it answers \qty{1}{\percent} of held-out questions (\num{147} of \num{150} entities score zero), confirming the films are outside its parametric knowledge and leaving substantial headroom for the adapter to fill.
The gain is therefore measured almost entirely on entities the base does not know, and we do not interpret the three-entity base-known cell.
On single-valued relations, adapters recover \num{+0.243} \ac{em} (\(\mathrm{CI}_{\qty{95}{\percent}} = [\num{+0.174}, \num{+0.319}]\)) over the near-blind base, as shown in~\cref{tab:encoding}: they learn, and generalize, facts they demonstrably did not know before.
The all-relations gain of \num{+0.127} is lower only because it pools in multi-valued relations that exact match scores at zero by construction: the single-valued figure is the fair measure of weight-level storage.

\begin{table}
  \centering
  \caption{
    \Acs{cb} \acs{em} gain over base on held-out questions.
    Values are averaged over \num{150} clean single-film entities, and gains are shown with their \qty{95}{\percent} \acsp{ci}.
    The \mbox{\(p\)-values} are computed using one-sample bootstrap test on paired differences based on \num{50000} resamples; \(^{*}p < \num{0.05}\), \(^{**}p < \num{0.01}\), \(^{***}p < \num{0.001}\).
  }\label{tab:encoding}

  \begin{tabular}{l *{2}{S[table-format=1.3]} S[table-format=+1.3] c S[table-format=<1.3]@{}l}
    \toprule
    Setting                 & {Base} & {Adapted} & {Gain} & \(\mathrm{CI}_{\qty{95}{\percent}}\) &\multicolumn{2}{c}{\(p\)} \\
    \midrule
    Single-valued relations & 0.007  & 0.250     & +0.243 & \([\num{+0.174}, \num{+0.319}]\)     & <0.001 & \(^{***}\) \\
    All relations           & 0.010  & 0.137     & +0.127 & \([\num{+0.083}, \num{+0.170}]\)     & <0.001 & \(^{***}\) \\
    \bottomrule
  \end{tabular}
\end{table}

\paragraph{Adapter specificity.}
The \ac{cb} gain could in principle reflect a generic output-conditioning effect\---any adapter making the model terser\---rather than entity-specific knowledge.
We make sure this is not the case by comparing the correct adapter against an unrelated (wrong-entity) adapter, an untrained random-weight adapter, and the base model.
Only the correct adapter helps, by \num{+0.283} \ac{em} over base (\(\mathrm{CI}_{\qty{95}{\percent}} = [\num{+0.183}, \num{+0.400}]\), \(p < \num{0.001}\)), as shown in~\cref{tab:oracle}: the stored knowledge is specific to the entity the adapter was trained on.
The two negative controls separate in an informative way.
An unrelated but trained adapter leaves performance exactly at base (gain \num{0.000}, \(p = \num{1.000}\)): it encodes real facts, just the wrong entity's, and does not change any outcome.
An untrained adapter with randomly initialized weights scores \num{0.000}, below base (gain \num{-0.017}, \(p < \num{0.001}\)), destroying even the residual pretraining knowledge the base model has for one entity.
The \ac{lora} structure therefore contributes nothing on its own; the recovered knowledge comes entirely from gradient updates on entity-specific data.
The wrong adapter never changes any outcome relative to base\---every paired difference is exactly zero, hence \(p = \num{1.000}\)\---because the base model is at floor on \ac{cb} \ac{qa} (\num{0.017} \ac{em}) and injecting an unrelated adapter does not move outputs enough to flip any answer.
That floor also means the wrong-adapter contrast has little power to detect a small positive effect, so its \mbox{\(p\)-value} should not be read as evidence of equivalence; the contrast that carries the claim is the own-adapter gain.

\begin{table}
  \centering
  \caption{
    \Acs{cb} \acsp{em} of different adapters, averaged over \num{30} queries and shown with their \qty{95}{\percent} \acs{ci}.
    The random-weight adapter is an untrained \ac{lora} with randomly initialized weights (same rank and target modules), averaged over five seeds; the wrong adapter is a trained adapter belonging to a different entity.
    The \mbox{\(p\)-values} are computed using one-sample bootstrap test on paired differences based on \num{50000} resamples; \(^{*}p < \num{0.05}\), \(^{**}p < \num{0.01}\), \(^{***}p < \num{0.001}\).
  }\label{tab:oracle}

  \begin{tabular}{l S[table-format=1.3] S[table-format=+1.3] c S[table-format=<1.3]@{}l}
    \toprule
    Injected adapter      & {Exact match} & {Gain} & \(\mathrm{CI}_{\qty{95}{\percent}}\)                 & \multicolumn{2}{c}{\(p\)} \\
    \midrule
    Base (no adapter)     & 0.017         & {---}  & {---}                                                & \multicolumn{2}{c}{---} \\
    Random-weight adapter & 0.000         & -0.017 & \([\num{-0.050}, \phantom{-}\num{0.000}]\)           & <0.001& \(^{***}\) \\
    Other (wrong) adapter & 0.017         & 0.000  & \([\phantom{-}\num{0.000}, \phantom{-}\num{0.000}]\) & 1.000 & \\
    Own (correct) adapter & 0.300         & +0.283 & \([\num{+0.183}, \num{+0.400}]\)                     & <0.001& \(^{***}\) \\
    \bottomrule
  \end{tabular}
\end{table}

\paragraph{Answer cardinality.}
Single-valued relations (a film has one director, one writer, one release year) and multi-valued relations (many tags, several cast members) behave differently under \ac{em}, and the difference is a property of the metric, not of what the adapter learned.
\Cref{tab:cardinality} shows the pattern: on multi-valued relations, \ac{em} is \num{0.000} because the model emits a valid answer inside verbose output that never exactly equals a single gold token, yet \ac{am} is \num{0.854} for tags and set recall is \num{0.793}.
We therefore restrict \ac{em} claims to single-valued relations and report set metrics for multi-valued ones, leaving set-valued parametric memory to future work.
As shown in~\cref{tab:relations}, among single-valued relations the gain is consistent: the \qty{95}{\percent} \acp{ci} for \enquote{director}, \enquote{writer}, and \enquote{release year} all exclude zero, while the sparser \enquote{genre} and \enquote{language} relations do not.

\begin{table}
  \centering
  \caption{
    Multi-valued relations under three \acs{cb} metrics.
    \Acs{em} registers zero while \acs{am} and set recall show the adapter produces valid answers; the zero is a scoring artefact, not a learning failure.
  }\label{tab:cardinality}

  \begin{tabular}{lS[table-format=1.3]S[table-format=1.3]S[table-format=1.3]}
    \toprule
    Relation       & {Exact match} & {Any match} & {Set recall} \\
    \midrule
    Has tags       & 0.000         & 0.854       & 0.793 \\
    Starred actors & 0.000         & 0.500       & 0.389 \\
    \bottomrule
  \end{tabular}
\end{table}

\begin{table}
  \centering
  \caption{
    \Acs{cb} \acs{em} gain over base for each single-valued relation.
    Values are computed over that relation's held-out questions in the \num{150}-entity bank and shown with its \qty{95}{\percent} \acs{ci}.
    The \mbox{\(p\)-values} are from a one-sample bootstrap test on paired differences based on \num{50000} resamples; \(^{*}p < \num{0.05}\), \(^{**}p < \num{0.01}\), \(^{***}p < \num{0.001}\).
  }\label{tab:relations}

  \begin{tabular}{l S[table-format=+1.3] c S[table-format=<1.3]@{}l}
    \toprule
    Relation     & {EM gain over base} & \(\mathrm{CI}_{\qty{95}{\percent}}\)       & \multicolumn{2}{c}{\(p\)} \\
    \midrule
    Director     & +0.274              & \([\num{+0.161}, \num{+0.387}]\)           & <0.001& \(^{***}\) \\
    Writer       & +0.269              & \([\num{+0.115}, \num{+0.462}]\)           & 0.001 & \(^{**}\) \\
    Release year & +0.196              & \([\num{+0.089}, \num{+0.321}]\)           & 0.001 & \(^{**}\) \\
    Genre        & +0.182              & \([\phantom{-}\num{0.000}, \num{+0.455}]\) & 0.218 & \\
    Language     & +0.091              & \([\num{-0.182}, \num{+0.364}]\)           & 0.769 & \\
    \bottomrule
  \end{tabular}
\end{table}

\paragraph{Verbalization format.}
A natural design question is whether the verbalization function matters for weight-level storage: a fact-dense template might plausibly store better than fluent prose, in which case parametric \ac{kg} memory would depend on a carefully engineered encoding.
On an earlier \num{120}-entity bank we trained matched adapters under both a templated and a natural-prose verbalization of the same subgraphs, holding the \ac{qa} pairs fixed so that only the training text differs, and compared \ac{cb} gains.
We find no detectable difference: the paired difference is \num{+0.019} (\(\mathrm{CI}_{\qty{95}{\percent}} = [\num{-0.013}, \num{+0.050}]\), \(p = \num{0.275}\)), and no relation subset with more than two entities reaches significance.
We report this as a null result rather than evidence for either format: at this scale, weight-level storage does not appear to require a carefully engineered verbalization, plausibly because a \ac{kg} subgraph is already fact-dense in either rendering.
This ablation uses a different bank from the \num{150}-entity set of our main results, so we treat it as indicative rather than conclusive.

\subsection{Retrieval}\label{sec:retrieval}

A parametric memory is only useful if the right adapter can be selected for a query without its subgraph in hand.
\Cref{sec:encoding} showed the correct adapter recovers real \ac{cb} knowledge; here we ask whether similarity to the query can select an adapter that answers it.
Because an entity resolver returns the entity's own adapter whenever it is in the bank, the informative test for similarity-based selection is whether it can find a functional substitute\---so we hold each query entity's own adapter out of the gallery and ask whether any other adapter it selects recovers the query's answers.
The answer is a clean negative for similarity-based selection, and it is the paper's second main finding: the knowledge stored in an adapter does not transfer to, and cannot be recovered from, a semantically neighbouring entity's adapter, so choosing an adapter by semantic proximity to the query is the wrong mechanism for entity-keyed parametric memory.
We are careful about the scope of this claim: it is a negative result about similarity-based routing, not about adapter selection in general.
A conventional entity-resolution step\---title or identifier matching, or lexical search over the entity index\---would route many MetaQA questions correctly, since the questions typically name their anchor film and the bank holds exactly one adapter per film; a deployed system would use such a resolver (\cref{sec:cost}).
Our point is narrower: the embedding and weight-space geometry that a parametric system might hope to retrieve with do not encode answer containment, and so cannot substitute for entity resolution.

\paragraph{Closed-book retrieval.}
The retrieval question is only meaningful in a \ac{cb} setting: with the subgraph in context, the base model reads the answer directly and every condition ties, so a retrieval difference can only appear when the adapter is the sole source of knowledge.
\Cref{tab:retrieval} evaluates \ac{cb} retrieval: for each query we retrieve an adapter by question-embedding distance, by \(\mathbf{\Delta W}\) Frobenius geometry, and at random, inject it with no context, and score the query's held-out questions.
The correct adapter beats the base model by \num{+0.283}, but no query-driven method finds a substitute that recovers any of that gain: embedding and \(\mathbf{\Delta W}\) retrieval both land exactly at base (\num{0.000} gain, \(p = \num{1.000}\)), and the two random baselines are indistinguishable from base as well (\(p = \num{0.778}\) and \(p = \num{0.533}\)).
The two query-driven methods are not merely both at chance but exactly equivalent to each other, their paired difference being identically \num{0.000}: weight-space geometry adds nothing over a sentence embedding for this task.
Crucially, this is not a failure of the retriever to identify similar entities.
Question embeddings identify semantically appropriate neighbours cleanly: each entity is its own nearest neighbour, and near neighbours are genuinely related films (e.g.\ two films by the same director).
The failure is that a neighbour's adapter, however semantically apt, does not contain the answer to the query entity's questions.
Knowledge is stored locally in each adapter and does not transfer.

\paragraph{Retrieval protocol.}
Of the \num{150} entities \num{30} serve as queries, and the remaining \num{120} form the gallery, so a query entity's own adapter is never among the candidates; the oracle row of \cref{tab:retrieval} injects that held-out adapter directly as an upper bound.
In the embedding condition, we encode the question text and the verbalized subgraph of each gallery entity with a sentence-embedding model and rank adapters by cosine distance.
In the \(\mathbf{\Delta W}\) condition, we rank gallery adapters by their normalized Frobenius distance, aggregated over the top layers, to the query entity's own trained \(\mathbf{\Delta W}\).
This condition uses the held-out adapter as the query representation, which a deployed system would not have: it is therefore not a deployable retrieval method but a test of whether an entity's weight-space neighbours carry transferable knowledge.
Because the query is represented by the very adapter that holds its answers, the setting favours weight geometry: a failure here is not an artefact of a weak query representation.
We report downstream \ac{cb} \ac{em} after injecting the top-ranked adapter, which is the quantity that matters for parametric memory: a retrieved adapter is only useful if it answers the question.
Rank statistics add little here, since the own adapter cannot be retrieved; the functional analogue\---the fraction of queries whose top-ranked adapter answers at least one held-out question\---is \num{0.020} for a randomly drawn gallery adapter, and \(\mathbf{\Delta W}\) retrieval reaches it for one query of \num{30}.
In that single case the base model already answers the question with no adapter, so the success reflects pretraining knowledge rather than transfer between adapters.
The claim therefore rests on the downstream \ac{em} gap between the oracle and every query-driven method.

\begin{table}
  \centering
  \caption{
    \Acs{cb} \acs{em} gain over base for query-driven adapter retrieval.
    Values are computed over \num{30} queries and shown with their \qty{95}{\percent} \acs{ci}; each query's own adapter is held out of the \num{120}-adapter gallery, so retrieval methods select among other entities' adapters.
    \enquote{Oracle} injects the held-out adapter as an upper bound.
    The \mbox{\(p\)-values} are computed using one-sample bootstrap test on paired differences based on \num{50000} resamples; \(^{*}p < \num{0.05}\), \(^{**}p < \num{0.01}\), \(^{***}p < \num{0.001}\).
  }\label{tab:retrieval}

  \begin{tabular}{l S[table-format=1.3] S[table-format=+1.3] c S[table-format=<1.3]@{}l}
    \toprule
    Retrieval method                & {Exact match} & {Gain} & \(\mathrm{CI}_{\qty{95}{\percent}}\)                 & \multicolumn{2}{c}{\(p\)} \\
    \midrule
    Base (no adapter)               & 0.017         & {---}  & {---}                                                & \multicolumn{2}{c}{---} \\
    Question embedding              & 0.017         & 0.000  & \([\phantom{-}\num{0.000}, \phantom{-}\num{0.000}]\) & 1.000 & \\
    \(\mathbf{\Delta W}\) Frobenius & 0.017         & 0.000  & \([\phantom{-}\num{0.000}, \phantom{-}\num{0.000}]\) & 1.000 & \\
    Random (global)                 & 0.010         & -0.007 & \([\num{-0.047}, \num{+0.023}]\)                     & 0.778 & \\
    Random (within-domain)          & 0.007         & -0.010 & \([\num{-0.050}, \num{+0.020}]\)                     & 0.533 & \\
    Oracle (own adapter)            & 0.300         & +0.283 & \([\num{+0.183}, \num{+0.400}]\)                     & <0.001& \(^{***}\) \\
    \bottomrule
  \end{tabular}
\end{table}

\paragraph{Geometry.}
On the clean bank, pairwise \(\mathbf{\Delta W}\) Frobenius distances do correlate with subgraph semantic distance: Spearman \(\rho\) is \num{+0.329} over all layers and \(\num{+0.352}\) over the top eight layers.
To confirm this is a real signal rather than an artefact of the distance formula, we use two null controls: permuting entity labels across adapters and column-shuffling each \(\mathbf{\Delta W}\) (destroying low-rank structure while preserving the marginal value distribution).
Both nulls stay at or below \(|\rho| = \num{0.157}\) over \num{1000} resamples, far below the observed value (\(p < \num{0.001}\)).
Consistent with the finding that later layers carry more of the entity-specific signal, the top-layer aggregation in~\cref{tab:geom-sem} selects layers \numlist{15;16;17;19;20;21;22;23}.
\Cref{fig:scatter} plots the correlation directly: the trend is real but weak, and the \num{30} pairs selected by \(\mathbf{\Delta W}\) retrieval are spread along it like any other pairs, with none adding knowledge over the base model.
Yet, this correlation does not translate into functional retrieval: \(\mathbf{\Delta W}\) retrieval performs exactly as embedding and random do.
Weight geometry thus encodes which entities are similar but not which adapter answers a given question\---a geometry--function dissociation, in which a real geometric signal carries no matching functional advantage.
We therefore read the \(\rho\)--retrieval gap not as a weak or noisy geometric signal, but as evidence that entity similarity and answer containment are genuinely different~relations.

\begin{table}
  \centering
  \caption{Geometry--semantics correlation on the clean bank, with null controls.}\label{tab:geom-sem}



  \caption{
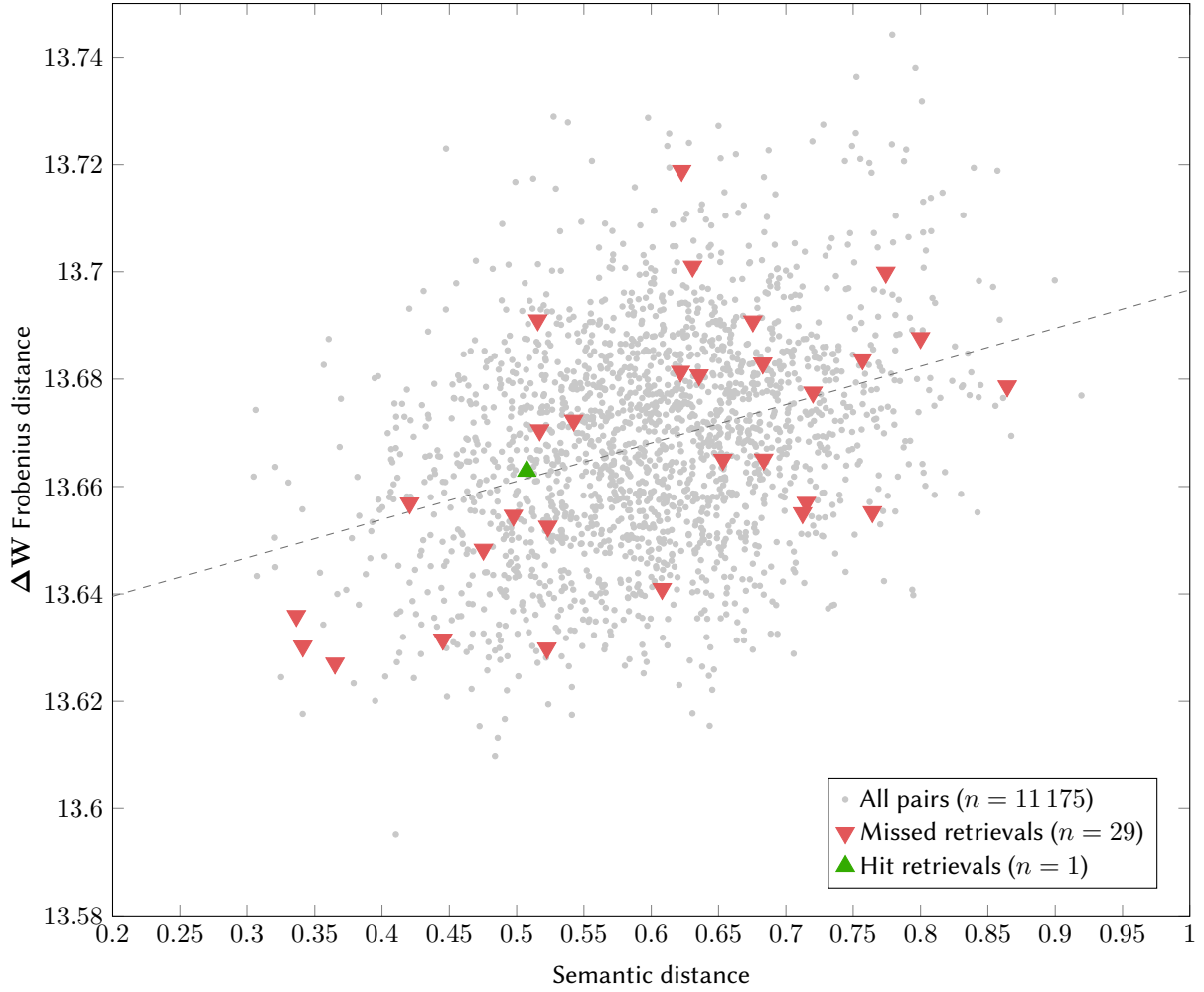
    Weight geometry versus semantic distance for all \num{11175} adapter pairs.
    The dashed line shows the correlation between \(\mathbf{\Delta W}\) Frobenius distance and semantic distance (\(\rho = \num{+0.329}\)); it should be noted that the effect is small in absolute terms (the \(Y\)-axis is zoomed to the observed range, about \qty{1.2}{\percent} of the norm), since all adapters share nearly identical weight magnitudes; highlighted are the \num{30} query--gallery pairs selected by \(\mathbf{\Delta W}\) retrieval, which are distributed like any other pair and none of which improves over the base model\---so proximity in weight space does not identify an adapter that answers the query.
  }\label{fig:scatter}
\end{figure}

\subsection{Composition}\label{sec:composition}

Multi-hop questions require combining knowledge across entities, and the natural mechanism is to merge the relevant adapters.
Subgraph adapters are, in principle, candidates for merging: each encodes a small, low-rank perturbation, and if per-entity perturbations occupy largely distinct directions, merging need not cause destructive interference.
The obstacle we encounter is not interference but the benchmark and the retrieval result above.
Concretely, MetaQA's two-hop questions ask about an intermediate entity (e.g.\ \enquote{what films share a director with this film?}), and their answers require knowledge contained in neither of the two entity subgraphs a merge would combine, nor in the two subgraphs placed jointly in context.
In our tests every condition, including the both-subgraphs-in-context oracle, scored at floor, so the experiment could not isolate a composition effect: the information needed was absent from the inputs by construction.
In addition, entity pairs were formed by a shared coarse attribute (genre, release year, or language) that rarely matches the relation the question asks about, so the second subgraph seldom bore on the question at all.
We therefore do not report a composition result and instead state the problem precisely, because it connects directly to the retrieval finding of~\cref{sec:retrieval}.
Both point to the same gap: knowledge is stored locally in per-entity adapters, and query-driven retrieval does not assemble the right knowledge for a multi-entity question.
Closing this gap\---learning which adapters to combine and how to weight them for a given query, rather than selecting by similarity\---is the central open problem that a learned composition mechanism would address and is left to future~work.

\section{Discussion}

\subsection{Interpreting the dissociation}

The paper's two negative results\---chance-level retrieval (\cref{sec:retrieval}) and the composition mismatch (\cref{sec:composition})\---share a single behavioural signature: an adapter's knowledge is available when the adapter is the correct one and absent otherwise, with no graceful degradation for near neighbours.
We take this seriously as a finding rather than a nuisance, and separate what we can and cannot conclude about its cause.

Three facts hold with statistical support.
First, the correct adapter carries entity-specific knowledge: it beats the base model by \num{+0.283} \ac{cb}, while an unrelated trained adapter leaves performance exactly at base and an untrained random-weight adapter falls below it.
Second, similarity does not imply transfer: the question-embedding retriever selects genuinely related neighbours (same-director or same-genre films), yet applying those neighbours' adapters recovers nothing over random.
Third, the failure is not a failure of the geometry to organize by semantics\---\(\mathbf{\Delta W}\) distances correlate with subgraph semantics at \(\rho = \num{+0.329}\), well above the permutation nulls (\(|\rho| \le \num{0.157}\))\---but a failure of that organization to predict answer containment.

Why do neighbouring adapters not help? We can point to at least two mechanisms our experiments do not distinguish.
One is \emph{answer non-overlap}: two similar films rarely share the specific object a question asks for (given two similar films, their directors are almost always different), so even perfect transfer would not help\---a property of the \ac{qa} task, not of the adapters.
The other is \emph{storage locality}: fine-tuning may write each entity's facts into a subspace a different entity's adapter does not activate, making knowledge mechanically inaccessible across adapters.
Both predict the same retrieval curve, and separating them would require shared-answer cases at a scale MetaQA's single domain does not~allow.

\subsection{Cost and deployment}\label{sec:cost}

Because query-driven adapter selection is unsolved (\cref{sec:retrieval}), the cost analysis carries the central justification for storing knowledge in weights rather than retrieving subgraph text.
The case rests on one property: weight injection consumes zero context tokens at query time, whereas graph \ac{rag} pays a per-query token cost that grows with subgraph size, as shown in~\cref{tab:cost}.
On MetaQA this cost is modest: measured with the Qwen3.5-2B tokenizer over all \num{150} clean entities, a serialized one-hop subgraph is a median of \num{60} tokens (the interquartile range is \([\num{53}, \num{70}]\)), because the benchmark's entities are sparse\---typically six or seven single-line fields.
The \num{60} tokens should therefore be read as a floor rather than a headline: the saving is small here precisely because MetaQA subgraphs are small, and it would scale up on the dense, multi-hop, property-rich subgraphs of industrial \acp{kg}, where in-context serialization runs to hundreds or thousands of tokens per query.
This zero-token property is the dimension on which parametric memory wins unconditionally on context cost.
We are careful not to overclaim it as a latency or energy win: injecting and swapping a per-query adapter is itself a cost that graph \ac{rag} does not pay, and we have not measured end-to-end latency or energy, so we claim the context-token saving only, not a net speedup.

\begin{table}
  \centering
  \small
  \caption{
    Cost comparison.
    The zero context-token cost of parametric memory is definitional, as no subgraph enters the prompt; token and byte figures are medians over the same \num{150} entities used throughout, measured on the verbalized subgraph text used for training (tokens with the Qwen3.5-2B tokenizer, bytes as UTF-8); the adapter size is the raw checkpoint, identical across all \num{150} adapters, and is a one-time cost amortized over all queries to an entity.
  }\label{tab:cost}

  \begin{tabular}{l c c}
    \toprule
                           & Graph RAG (in-context) & Parametric (ours) \\
    \midrule
    Context tokens / query & \num{60}               & \num{0} \\
    Bytes shipped / query  & \qty{188}{\byte}       & \qty{10.4}{\mebi\byte} (one-time) \\
    \bottomrule
  \end{tabular}
\end{table}

That token saving comes at a one-time storage cost.
An adapter is far larger than the subgraph it encodes (\qty{10.4}{\mebi\byte} of weights versus \qty{188}{\byte} of serialized triples, a ratio of roughly \num{58000} times more) so, on bytes alone, parametric memory is favourable only when an entity is queried often enough to amortize its adapter, or when the binding constraint is context budget rather than storage.
That ratio also bounds scalability honestly: a bank covering one million entities would occupy on the order of \qty{10}{\tebi\byte} of checkpoints, plus a one-time offline training cost per entity, so the approach suits stable, frequently queried subgraphs rather than web-scale or rapidly changing graphs.
At query time the parametric layer injects weights and never ships the raw subgraph, which reduces raw-context transmission when the graph is proprietary.
We frame this as reduced raw-context exposure rather than leakage resistance: our own positive result shows that facts are recoverable from the weights, so the adapter is itself a potential leakage surface, and we do not test resistance to extraction.
Both benefits are real only for the storage part of the system: because selection is unsolved (\cref{sec:retrieval}), a deployed system would today pair adapter storage with a conventional question-to-entity resolver rather than weight-space retrieval.

Parametric \ac{kg} memory is not free.
It pays a one-time offline encoding cost and, once compiled, a fact update requires re-encoding the affected adapter.
For graphs that change continuously, or for one-shot queries over a graph that is never reused, in-context graph \ac{rag} remains the better trade-off.
Parametric memory pays off when a stable subgraph is queried many times, when context budget is the binding constraint, or when shipping raw subgraphs is undesirable.

\section{Related work}

\paragraph{Parametric RAG and what LoRA encodes.}
A line of work encodes retrieved documents directly into model parameters rather than into context: Parametric \ac{rag}~\citep{su2025parametric} fine-tunes a \ac{lora} adapter per document and averages retrieved adapters at inference, \ac{dyprag}~\citep{tan2025dynamic} trains a hypernetwork to map documents to adapters on the fly, and Poly-PRAG~\citep{su2025using} learns a shared adapter basis with per-document routing.
These systems target document-level, single-hop \ac{rag} and implicitly assume that document-encoded adapters carry the document's knowledge\---which presumes that fine-tuning stores facts in the weights rather than only teaching the model to exploit in-context text.
We apply the parametric-memory idea to \ac{kg} subgraphs and test that assumption directly, using closed-book evaluation to isolate the weight-resident component and an own-versus-other-adapter contrast to confirm the recovered knowledge is entity-specific: subgraph adapters do encode recoverable \ac{cb} knowledge, but it cannot be retrieved from the query alone.

\paragraph{KG question answering, graph RAG, and merging.}
Standard \ac{qa} on \ac{kg} and graph \ac{rag} approaches retrieve and serialize subgraphs into the prompt; we do not replace retrieval but move the retrieved knowledge from the context window into the weights, trading a recurring per-query token cost for a one-time offline encoding cost, and use an in-context oracle as an upper bound.
For composition, task arithmetic~\citep{ilharco2023editing} and \textsc{ties}-merging~\citep{yadav2023ties} compose weight perturbations and study interference; whether per-entity subgraph adapters can be merged without destroying each other's knowledge, and whether a merged adapter supports reasoning that spans the entities it combines, is the open question our composition experiment probes (\cref{sec:composition}).

\section{Conclusion}

We framed a knowledge graph as a parametric memory: a bank of \ac{lora} adapters, one per film, queried by weight injection with no subgraph text in context.
We found that subgraph-trained adapters encode context-free, \ac{cb}-recoverable knowledge that generalizes to unseen questions\---a \num{+0.243} \ac{em} gain on single-valued relations over a near-blind base.
Only the correct adapter recovers this knowledge (an oracle gap of \num{+0.283} over base), but no query-driven method finds another adapter that substitutes for it: embedding and \(\mathbf{\Delta W}\) geometry retrieval both perform at chance, because knowledge is stored locally and does not transfer to semantically neighbouring adapters.

These findings come with scope conditions.
This is a single-benchmark, single-model, workshop-scope study with model-generated \ac{qa} pairs.
The headline \ac{cb} claim is restricted to single-valued relations, since multi-valued ones (e.g.\ tags or cast) require a set-valued notion of parametric memory we leave to future work.
Our single-domain (film) entities leave open whether the retrieval negative persists across broad domains, where \(\mathbf{\Delta W}\) geometry separates more strongly.
Within these bounds, storing knowledge in weights is feasible at zero query-time token cost.
The open problem is not storage but composition: learning which local adapters to combine for a multi-entity~query.

\section*{Declaration on Generative AI}
During the preparation of this work, the authors used Claude Opus 4.8 and Claude Opus 5 in order to: draft content, paraphrase and reword, improve writing style, draft abstract, check grammar and spelling, simulate peer review, and enhance content.
After using this service, the authors reviewed and edited the content as needed and take full responsibility for the publication's content.

\bibliography{storage-retrieval-gap}

\appendix

\section{Mathematical details}\label{sec:math}

\subsection{SVD representation}

A \ac{lora} adapter modifies each weight matrix \(\mathbf{W}_0 \in \mathbb{R}^{d_\mathrm{out} \times d_\mathrm{in}}\) with a low-rank perturbation
\begin{equation*}
  \mathbf{W} = \mathbf{W}_0 + \mathbf{\Delta W} = \mathbf{W}_0 + \mathbf{B} \mathbf{A}
\end{equation*}
where \(\mathbf{B} \in \mathbb{R}^{d_\mathrm{out} \times r}\) and \(\mathbf{A} \in \mathbb{R}^{r \times d_\mathrm{in}}\).
We work exclusively with \(\mathbf{\Delta W} = \mathbf{B} \mathbf{A}\) rather than the raw factors.
For each layer--module pair, we compute the truncated \ac{svd}
\begin{equation*}
  \mathbf{\Delta W} = \mathbf{U} \operatorname{diag}(\boldsymbol{\sigma}) \mathbf{V}^\top
\end{equation*}
where \(\mathbf{U} \in \mathbb{R}^{d_\mathrm{out} \times r}\), \(\boldsymbol{\sigma} \in \mathbb{R}^r\) the vector of singular values, and \(\mathbf{V}^\top \in \mathbb{R}^{r \times d_\mathrm{in}}\).
Only \(\mathbf{U}\), \(\boldsymbol{\sigma}\), and \(\mathbf{V}^\top\) are stored: the full \(d_\mathrm{out} \times d_\mathrm{in}\) matrix is never formed.
Storing the rank-\(r\) factors rather than the dense \(\mathbf{\Delta W}\) for every layer--module pair reduces analysis-time memory by roughly the same \(d/r\) factor, turning what would be several gibibytes per adapter into a few tens of mebibytes; the on-disk checkpoint, which stores only the \ac{lora} factors, is smaller still at \qty{10.4}{\mebi\byte}.

The \ac{svd} is computed efficiently via QR decomposition of the low-rank factors, in \(\operatorname{\mathcal{O}}(d r^2)\) time rather than \(\operatorname{\mathcal{O}}(d^3)\) for a direct \ac{svd}.
We first take QR factorizations \(\mathbf{B} = \mathbf{Q}_B \mathbf{R}_B\) and \(\mathbf{A}^\top = \mathbf{Q}_A \mathbf{R}_A\), then \ac{svd} the small product:
\begin{equation*}
  \mathbf{U}, \boldsymbol{\sigma}, \mathbf{V}^\top = \operatorname{SVD}(\mathbf{R}_B \mathbf{R}_A^\top)
\end{equation*}
and finally project back via \(\mathbf{U} \leftarrow \mathbf{Q}_B \mathbf{U}\) and \(\mathbf{V}^\top \leftarrow \mathbf{V}^\top \mathbf{Q}_A^\top\).
The normalized representation \(\mathbf{\Delta\tilde{W}} = \mathbf{\Delta W} / \lVert \mathbf{\Delta W} \rVert_F\) is captured implicitly by \(\tilde{\boldsymbol{\sigma}} = \boldsymbol{\sigma} / \lVert \boldsymbol{\sigma} \rVert_2\), since \(\lVert \mathbf{\Delta W} \rVert_F = \lVert \boldsymbol{\sigma} \rVert_2\).

\subsection{Frobenius distance}

For a pair of adapters \((i, j)\) at a given layer--module pair \(\ell \in \mathcal{L}\), the normalized Frobenius distance is
\begin{equation*}
  \begin{split}
    d_\ell(i, j) & = \left\lVert \mathbf{\Delta \tilde{W}}_i^\ell - \mathbf{\Delta \tilde{W}}_j^\ell \right\rVert_F \\
    & = \sqrt{2 - 2 \left\langle \mathbf{\Delta \tilde{W}}_i^\ell, \mathbf{\Delta \tilde{W}}_j^\ell \right\rangle_F}
  \end{split}
\end{equation*}
using the identity
\begin{equation*}
  \lVert A - B \rVert_F^2 = \lVert A \rVert_F^2 + \lVert B \rVert_F^2 - 2 \langle A, B \rangle_F
\end{equation*}
together with \(\lVert \mathbf{\Delta \tilde{W}} \rVert_F = 1\).
The inner product is evaluated directly from the \ac{svd} factors, without ever materializing \(\mathbf{\Delta W}\):
\begin{equation*}
  \left\langle \mathbf{\Delta \tilde{W}}_1, \mathbf{\Delta \tilde{W}}_2 \right\rangle_F = \tilde{\boldsymbol{\sigma}}_1^\top \left(\mathbf{M}_U \odot \mathbf{M}_V\right) \tilde{\boldsymbol{\sigma}}_2
\end{equation*}
where \(\mathbf{M}_U = \mathbf{U}_1^\top \mathbf{U}_2\) and \(\mathbf{M}_V = \mathbf{V}_1^\top \mathbf{V}_2\) are both in \(\mathbb{R}^{r \times r}\), and \(\odot\) denotes the Hadamard (element-wise) product.
The order of factors in \(\mathbf{M}_V\) matters: \(\mathbf{V}_1^\top \mathbf{V}_2\) is \(r \times r\), whereas \(\mathbf{V}_1 \mathbf{V}_2^\top\) would be \(d_\mathrm{in} \times d_\mathrm{in}\) and defeat the purpose.
The total cost is \(\operatorname{\mathcal{O}}(r^2 (d_\mathrm{out} + d_\mathrm{in}))\) per adapter pair, roughly \num{1000} times cheaper than the dense computation, which for \(r = \num{8}\) is two to three orders of magnitude at the layer widths of current models.

The aggregated distance is computed as the Euclidean norm over per-layer distances as
\begin{equation*}
  d(i, j) = \sqrt{\sum_{\ell \in \mathcal{L}} {d_\ell(i, j)}^2}
\end{equation*}
where \(\mathcal{L}\) is the set of selected layer--module pairs.
Each layer is normalized independently before aggregation, so that layers with larger weight magnitudes do not dominate the aggregate.

\section{Prompt template}\label{app:prompts}

We report the exact template used for training and evaluation.
The \texttt{\{context\}} field is the verbalized entity document; for the \ac{cb} condition it is the empty string.
The \texttt{\{ground\_truth\_answer\}} field is used only for training.

\begin{prompt}
You are a factual question-answering assistant. Answer the question below using only the information provided. Be concise and accurate.

Context: {context}

Question: {question}

Answer: {ground_truth_answer}
\end{prompt}

\end{document}